# Spatial Modeling of Oil Exploration Areas Using Neural Networks and ANFIS in GIS

Misagh, N. *[1,] Ashouri, M. *[2], neisany samany, N. [3] A. Kakroodi, A. [4] Alavipanah,, S.k . [5]

1- Nouraddin Misagh[*], Department of Remote Sensing and GIS, College of Geography, University of Tehran

2- Mohammadreza Ashouri, Department of Algorithms and Computation, Collage of Engineering, University of Tehran

3- Najmeh neisany samany , Assistant professor in Department of Remote Sensing and GIS, College of Geography, University of Tehran

4- Ataollah Abdollahi Kakroodi, Assistant professor in Department of Remote Sensing and GIS, College of Geography, University of Tehran

5- Seyed Kazem Alavipanah , professor in Department of Remote Sensing and GIS, College of Geography, University of Tehran

## Abstract

Exploration of hydrocarbon resources is a highly complicated and expensive process where various geological, geochemical and geophysical factors are developed then combined together. It is highly significant how to design the seismic data acquisition survey and locate the exploratory wells since incorrect or imprecise locations lead to waste of time and money during the operation. The objective of this study is to locate high-potential oil and gas field in 1: 250,000 sheet of Ahwaz including 20 oil fields to reduce both time and costs in exploration and production processes. In this regard, 17 maps were developed using GIS functions for factors including: minimum and maximum of total organic carbon (TOC), yield potential for hydrocarbons production (PP), $T_{max}$ peak, production index (PI), oxygen index (OI), hydrogen index (HI) as well as presence or proximity to high residual Bouguer gravity anomalies, proximity to anticline axis and faults, topography and curvature maps obtained from Asmari Formation subsurface contours. To model and to integrate maps, this study employed artificial neural network and adaptive neuro-fuzzy inference system (ANFIS) methods. The results obtained from model validation demonstrated that the 17×10×5 neural network with R=0.8948, RMS=0.0267, and kappa=0.9079 can be trained better than other models such as ANFIS and predicts the potential areas more accurately. However, this method failed to predict some oil fields and wrongly predict some areas as potential zones.

1. Introduction

Although identifying and locating factors such as source rocks, reservoir rocks, and cap rocks (oil traps) are basis of exploration, there are still many problems in various fields with no solution. This is mainly due to no direct access to oil reserves in depths. Since source rocks are located deep underground, we cannot certainly locate oil fields in a region. As a result, it is always probable to face a dry well after spending over 100 million dollars to drill an exploration well (Bott & Carson, 2007). Thus, selecting the best possible path for seismic data acquisition which is highly costly and determining the best location for drilling exploration wells are of particular importance since incorrect or careless positioning imposes large costs or may cause serious problems for the exploration project. The main issue is therefore how to determine the location of potential areas within the target zone in less possible time and with minimum possible exploration and production costs. Fig. 1 shows the exploration process of an oilfield and the costs.

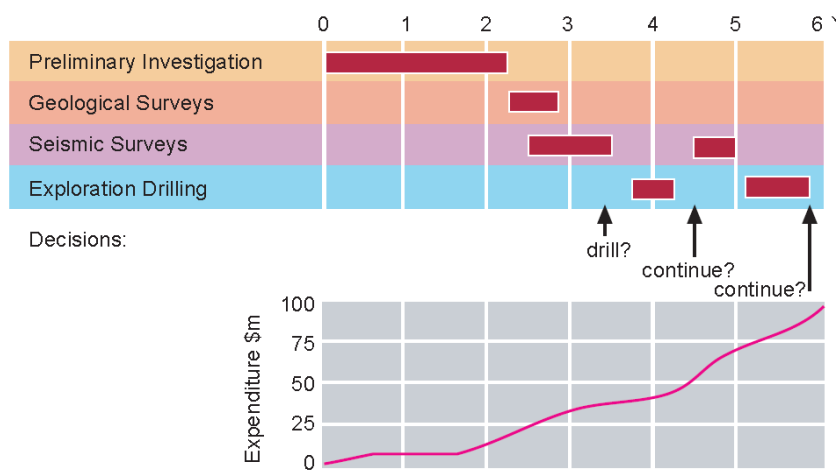

Fig. 1: Process of an exploration program and the costs (Frank, et al, 2008)

In exploration or development process of an oilfield, various geological, geochemical and geophysical factors are developed and combined together. Today, potential-positioning procedures are usually done through GIS environment (Carranza, 2008). Geographic Information System (GIS) allows gathering, storing, retrieving, managing, processing and displaying spatial and descriptive data to support decisions made based on spatial data. GIS generates new information by combining different data layers using different methods with different views (Malczewski, 1999). Application of GIS in oil exploration is almost new. Using a set of analytical tools, an efficient GIS can integrate data and make it possible for users to overlap and change data sets as a map to analyze the potential or development of existing fields. This would reduce exploration time and cost. In general, data integration models in GIS are divided into two general categories as data-driven and knowledge-driven models (Bonham Carter 1994, Nikravesh, et al, 2003). Data-driven models include neural network, weight of evidence, and logistic regression and knowledge-driven models include multi-criteria evaluation, fuzzy logic, Dempster-Shafer, and index overlay (Bonham Carter, 1994). To map the potential, data-driven models act based on exploration and empirical data and calculate weights using statistical methods for data gathered in the target zone. Data-driven models can be developed in areas where there are enough training data. On the other hand, knowledge-driven models are developed based on expert estimation and inferential data and are used in unknown areas.

As mentioned, a data-driven method is artificial neural network (ANN) algorithm. ANNs are powerful tools both in pattern recognition and solving complex natural problems (Frate, et al, 2004). Even in cases of dependency of input variables and the presence of noise in data, the accuracy of ANN is acceptable. ANN is highly flexible with the ability of retraining by new input data. By managing and processing huge input data sets, ANN is able to correctly analyze the relationship between data sets and to extract evidence for pattern recognition objectives. As a result, ANN can be employed as a reliable method to define the potential oil fields. Furthermore, hybrid intelligent systems combining fuzzy logic and neural networks allow taking advantages of both. Fuzzy logic does the reasoning under uncertainty, while ANN includes learning, compliance, and parallel-distributed processing. A fuzzy inference system takes advantages of neural networks to overcome the limitations of fuzzy logic. A main advantage of fuzzy inference system is taking advantage of the learning ability of neural networks to prevent the costly and time consuming process of rule development by inference engine based on the logic fuzzy. It can also take advantages of diverse data types (numerical, logical, linguistic variable, etc.), managing inaccurate, incomplete, and ambiguous information, learning ability, and imitating the decision-making process (Jain S, Khare M, 2010).

Zargani et al (2003) used GIS and weight of evidence to study high potential oil-fields in Libya. In this study, they initially estimated the probability of hydrocarbon presence based on previous explorations and updated it by integrating geological factors which confirmed the hydrocarbon presence (Zargani, et al, 2003). In her MS thesis entitled "Oil Exploration using GIS-based Fuzzy Logic Analysis", Lisa Bingam employed fuzzy logic to develop a route map for oil exploration considering geological, seismic, and economic factors. After model validation through existing oil fields in sedimentary zones at north of South America, she suggested this model to map potential oilfields in other parts of the world (Bingam, et al, 2011). Attila developed a 3D subsurface map in Turkey using gravity data. He then used the map along with geochemical indicators such as $S_1$ and $S_2$ peak, total organic carbon, and $T_{max}$ to interpret subsurface hydrocarbon reservoirs in Tuz Gölü basin, Turkey (Attila, 2008).

The main objective of this study is to design and implement a GIS-based model employing a hybrid intelligent system (combining fuzzy logic and neural network) to locate high-potential hydrocarbon zones for further exploration operations including seismic survey and exploration drilling. Based on the experience of major oil companies such as Shell in application of GIS, the current study can be considered as a starting point for further researches proving the capability of GIS in problem solving at petroleum industry.

## 2. Research Methodology

The research process can be summarized as following steps:

1. Defining required standards and gathering useful data for exploration process at potential oil fields including geological, magnetic, gravimetric, and seismic maps.
2. Creating new layers and factor maps for oil fields using raw data as a database.
3. Creating oil-fields maps for the study area to be used in neural network and ANFIS models as objective data.
4. Converting factor maps to ASCII format and inserting them in MATLAB.
5. Designing MLP neural network structure and ANFIS model.
6. Selecting the best model based on validation indicators in order to explore potential oil-fields.
7. Developing an oil-potential map for the study area using the designed models.

### 2.1. General Characteristics of Study Area

The study area is 1: 250,000 sheet of Ahwaz, southwest of Iran. Most part of the area is located in Dezful embayment. Fig. 2 shows the study area with the discovered oil fields. This area is approximately 60,000 $Km^2$ wide containing 8% of the world's proven oil fields and 15% of the world's gas filed (Sherkati et al., 2004). The main source rock is Kajdomi Formation deposited at anaerobic marine sediments which is extended more in Fars and Dezful than Lorestan. With a high content of organic carbon, Kajdomi is considered as a main source rock in Zagros having a high oil potential so that it acts as the source rock for Asmari and Sarvak reservoir rocks in Dezful embayment. It is worth mentioning that Zagros Basin is considered as one of the most important oil basins in the world, so it has always been a target in Iran oil industry experiencing about a hundred years of oil operations (Ashkan, 2004).

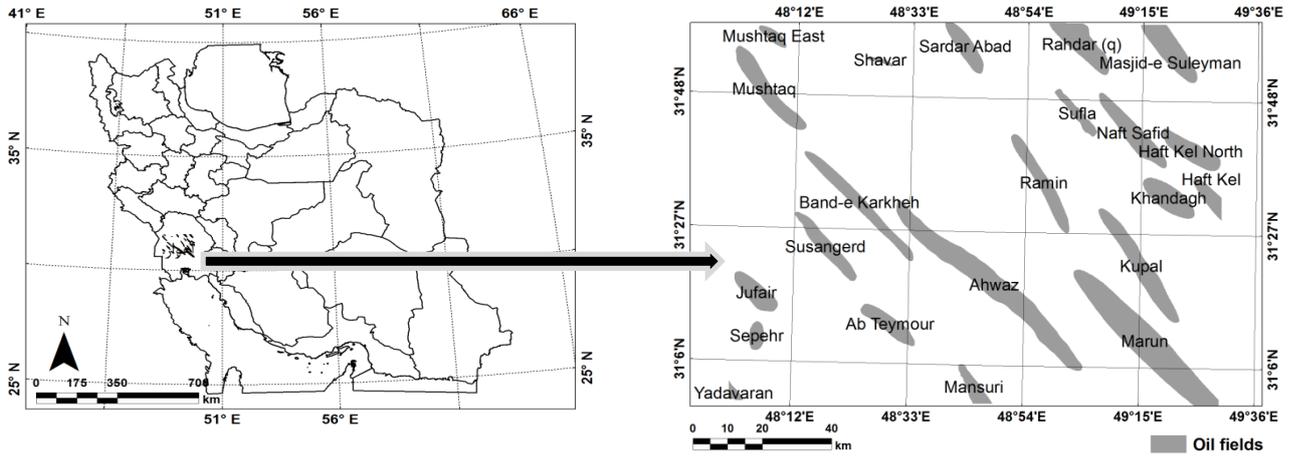

Fig. 2: The study area with oil fields

ANNs are parallel distributed processing models that can recognize very complex patterns in data sets (Mohaghegh & Ameri, 1995). One of the simplest, yet most effective arrangement proposed for use in modeling real neural models is multilayer perceptron model (MLP) consisting of an input layer, one or more hidden layers and an output layer (Menhaj, 1999) (Fig. 3). In neural networks, neurons are main elements, and weights are the connection basis. After using input data to train the system, outputs are calculated based on the minimum error. Here, error is defined the difference between calculated output and input signal. A common approach to reduce the model error is the back-propagation algorithm reducing the gradient.

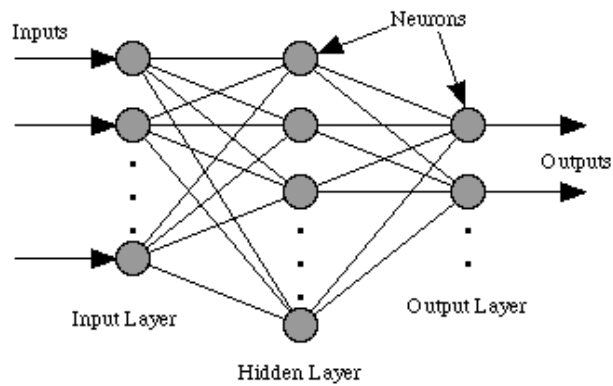

Fig 3: A view of a Multilayer Perceptron network

MLP procedure is as follows:

Suppose $t_{pj}$ is the desired output for pattern $p$ at node/neuron $j$ and $o_{pj}$ is the actual output of $j$. Here, $w_{pj}$ is the weighting factor for the connection line from neuron $i$ to neuron $j$.

1. Initial values for weighting factors and thresholds are selected.
2. All weights and thresholds are set to small random figures.
3. Desirable inputs and outputs are introduced to the network.

Here, $X_P = x_0, x_1, x_2, \ldots x_{n-1}$ and $T_P = t_0, t_1, t_2, \ldots t_{m1}$ are introduced to the network as input and target output, respectively, where n denotes the number of input elements and m is the number of output elements. The weighting factor, $w_0$, is set equal to the negative of threshold value, $-\theta$, and $x_0$ is set to 1. In classification problems, all elements of $T_P$ are set equal to zero except the one representing the classification which is 1 and includes $X_P$.

4. The output is computed.

Each layer calculates $y_{pj}$ and sends it to the next layer.

$$y_{pj} = f\left[\sum_{i=0}^{n-1} w_i x_i\right] \quad (1)$$

The output of each unit, $j$, is obtained by applying the threshold function, $f$, on the weighted sum of inputs for that unit. For single-layer Perceptron, this function is a step function while for MLP, it is a Sigmoid function.

5. Weighting factors are adjusted. We begin from the outer layer and move back.

$$w_{ij}(t+1) = w_{ij}(t) + \eta \delta_{pj} o_{pj} \quad (2)$$

where $w_{ij}(t)$ represents weighting factors for the link from node $i$ to node $j$ at time $t$, η denotes gain coefficient and $\delta_{pj}$ shows the error of $p$ at $j$. Applying a gain coefficient (or learning coefficient) smaller than 1 (i.e. 0.1-0.9) on Eq. 2 makes the weighting factor changing rate slower, thus the network approaches the solution in shorter steps.

$$\delta_{pj} = k o_{pj}(1 - o_{pj})(t_{pj} - o_{pj}) \quad (3)$$

Eq. 3 is useful for the outer layer units, because both target output and actual output are known. Thus, it is not suitable for the hidden layer units where the target output is unknown. As a result, the error for units in the hidden layer is calculated as:

$$\delta_{pj} = k o_{pj}(1 - o_{pj}) \sum_{k} \delta_{pk} w_{jk} \qquad (4)$$

While the summation operates for the *k* units in the layer next to *j*.

### 2.2. Adaptive Neuro-Fuzzy Inference System (ANFIS)

ANFIS is a hybrid system combining adaptive neural networks and fuzzy inference systems developed by Young (1990, 1993). Similar to fuzzy model, ANFIS uses the empirical knowledge, and it can be trained similar to the neural network. Using hybrid learning process, ANFIS parameters can be set based on input and out data for modeling objectives (Jyh-Shing, Roger Jang, 1993). Totally, ANFIS system consists of 5 layers where each layer consists of some input variables and each input has two or more membership functions (Jang, 1993) (Fig. 4). To explain the ANFIS algorithm, we assume Sugeno fuzzy *If - Then* rules with two inputs (x and y) and one output z:

Rule 1:   If   $x = A_1$   and   $y = B_1$   then   $f_1 = (p_1 x + p_1 y + r_1)$

Rule 2:   If   $x = A_2$   and   $y = B_2$   then   $f_1 = (p_2 x + p_2 y + r_2)$

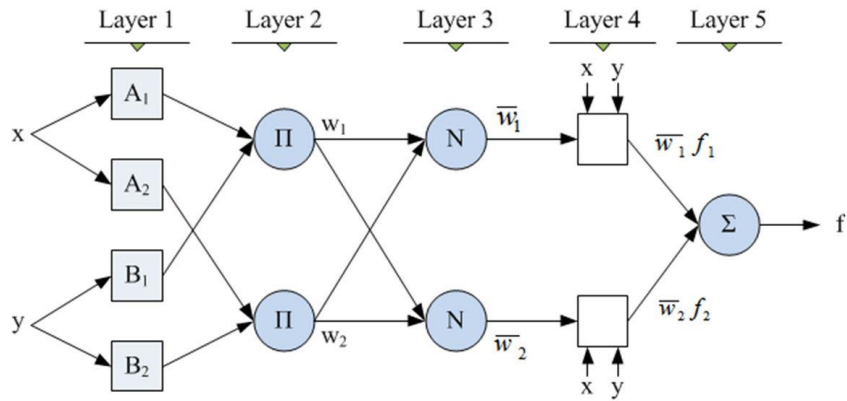

Fig 4: A five-layer ANFIS structure with two inputs and Sugeno-based rules (Jang, 1993)

This is a five-layer structure. As you can see, nodes of a certain layer have same functions. The output node *i* in layer *1* is labeled as $O_{1,i}$.

Layer 1: Each node within this layer consists an adaptive node with a node function. According to Eq. 5, we have:

$$O_{1,}i = \mu A_i(x) \text{ for } i = 1,2 \text{ or, } O_{1,}i = \mu B_i(y) \text{ for } i = 3,4 \tag{5}$$

Where *x* or *y* are the input for node i and A$_i$ or B$_i$ are the corresponding fuzzy set. In other words, the output of this layer is the membership value. Any suitable function which is continuously or segmentally differentiable (such as Gaussian, trapezoidal, and triangle functions) can be used as a membership function.

Layer 2: Each node in this layer applied the "and" fuzzy function and is named Π. The second layer nodes that are considered constant multiply the input signals and send the resulting output to the network representing firing strength of each rule (Eq. 6).

$$O_{2,}i = \mu A_i(x) \cdot \mu B_i(y) \text{ for } i = 3,4 \tag{6}$$

Layer 3: The nodes in layer 3 labeled *n* in Fig, 4, calculate the normalized weigh of rules (Eq. 7).

$$O_{3,}i = \frac{\overline{w}_i}{\sum_i w_i}, i = 1,2 \tag{7}$$

where $w_i$ is weight of rule i and $O_{3,}i$ is the normalized weight.

Layer 4: Each node in this layer is associated with a node function. Nodes here multiply the normalized weight of each fuzzy rule in the output of the rule posterior section (Eq. 8).

$$O_{4,}i = \overline{w}_i f_i = \overline{w}_i(p_i x + p_i y + r_i) \, i = 1,2 \tag{8}$$

Where $O_{4,}i$ is the output value of $i^{th}$ fuzzy IF-THEN rule, $w_i$ is the normalized weight of the third layer, and $(p_i x + p_i y + r_i)$ represents parameters for node *i*. The parameters of this layer are called "resulted parameters".

Layer 5: The only node in this layer is named Σ which calculate the summation of input signals and sends it to the output, the purpose of this layer is to minimize the difference between the network output and actual output (Jang, 1996). The total output of ANFIS is calculated as:

$$O_{5,}i = \sum_i \overline{w}_i f_i = \frac{w_i f_i}{\sum_i w_i} \quad i = 1,2 \tag{9}$$

where $O_5, i$ is the output of $i^{th}$ node in the fifth layer.

### 2.2.1. Hybrid Learning Algorithm

Hybrid learning algorithm is a combination of least squares and gradient-descent methods and includes two alternating phases. The reduced gradient returns error signals from the output layer to the input layer. This phase corrects the model parameters at prior section (membership functions). Total least square error corrects the model parameters at posterior section (coefficients of the linear relationship). In each round of training, the output of nodes are calculated as normal moving forward by layer 4. After calculating the error value, it is back-propagated to the inputs using gradient-descent algorithms, thus parameters are corrected (Jang, 1996). In ANFIS, rules are constant and what is optimized is the shape of membership functions. The total output $F$ of Fig. 4 is calculated using Eq. 10 as:

$$F = \frac{w1}{w1+w2}f1 + \frac{w1}{w1+w2}f2 \qquad (10)$$
$$= \overline{w1}f1 + \overline{w2}f2$$
$$= (\overline{w1}x)p1 + (\overline{w1}y)q1 + (\overline{w1})r1 + (\overline{w2}x)p2 + (\overline{w2}y)q2 + (\overline{w2})r2$$

Eq. 10 indicates that the output is a linear combination of parameters *r1, q1, p1 r2, q2, p2*. Thus, when the default parameters are constant, the total output is the linear combination of output parameters and the hybrid learning algorithm will act (Jang, 1996).

### 2.3. Factors Affecting Oil Fields Potential
### 2.3.1. Geochemical Data

Indicators used to determine the quantity and quality of the source rock includes total organic carbon (TOC), production potential (PP), $T_{max}$ peak, production index (PI), oxygen index (OI), hydrogen index (HI).

Oxygenation Index (OI): shows the oxygen content in kerogen. OI allows to evaluate the degree of oxidation of organic matters in source rocks, and there is direct relationship between OI and oxidation degree. OI is represented in milligram $CO_2$ per 1 gram of total organic carbon.

$$OI = {S_3}/{TOC} \quad mgCO_2/g \text{ TOC} \qquad (11)$$

Production Index (PI): This factor shows the level of thermal maturity of organic matter in source rocks. As thermal maturity increases, PI also increases. The migration factor has also a direct effect on PI. At the beginning of production phase, PI ranges from 0.05 to 0.1 and at the end of oil production, it reaches 0.3-0.4. PI is calculated as follows:

$$PI = \frac{S_1}{S_1 + S_2} \quad mgHC/g \text{ rock} \tag{12}$$

Production Potential (PP): This index is sum of $S_1$ and $S_2$ showing the hydrocarbon-production potential at source rock.

$$PP = S_1 + S_2 \quad mgHC/g \text{ rock} \tag{13}$$

Hydrogen Index (HI): This index shows the petroleum potential of the bed-rock. Higher HIs represents higher potential of source rocks (Kamali & Ghorbani, 2006). HI can be calculated as:

$$HI = S_2/TOC \quad mgHC/g \text{ TOC} \tag{14}$$

### 2.3.2. Subsurface Data Analysis

Structure-contour map is a map similar to a topographic map but for underground features like a formation boundary or an anticline. Some maps such as closure, roughness, and curvature maps can be developed based on structure-contour maps. Such maps show the position and shape of underground features.

### 2.3.2.1. Modeling the Anticline Axis

Since most discovered oil and gas fields in the world are located in anticline structures, proximity to an anticline can be an important factor in potential maps. The orientation of an anticline can be defined based on geological surface and sub-surface maps. However, the exposed part of an anticline at Earth's surface is in fact a small part of the anticline structure. In such cases, the continuity of sub-surface and exposed parts may be distorted by factors such as faults. On the other hand, there are many anticlines with no outcrop (McQuillin et al, 1984). Therefore, to have a real model for anticlines, we must consider all anticlines not only the ones mapped in geological maps. The terrain ruggedness index (TRI) is calculated as follows (Riley et al, 1999):

$$TRI = Y \left[ \sum (x_{ij} - x_{00})^2 \right]^{1/2} \tag{15}$$

where $x_{ij}$ is the height of each neighboring cell.

### 2.3.2.2. Curvature Map

Most structural traps (e.g. anticlines) are convex. To develop a map of potential traps in a region, in addition to a topographic map, a curvature map is also required. The surface curvature is related to the second derivative of the height. Curvature helps to identify the breaks of slope and roughness potential (Grohmann et al, 2011). Curvature can be imagined as a curve covered the anticline surface. The second derivative of this surface gives the inflection point of the function. Values higher than the inflection point will be considered as anticline and lower values as syncline. Curvature can be used as an index along with surface roughness to identify anticlines and synclines. Fig. 5 shows profiles obtained by applying the curvature index on a DEM. Since the values of subsurface contours are reverse of surface contours, the interpretation of results obtained by the curvature index is reverse of DEM, i.e. positive values represent anticlines and negative values represent synclines.

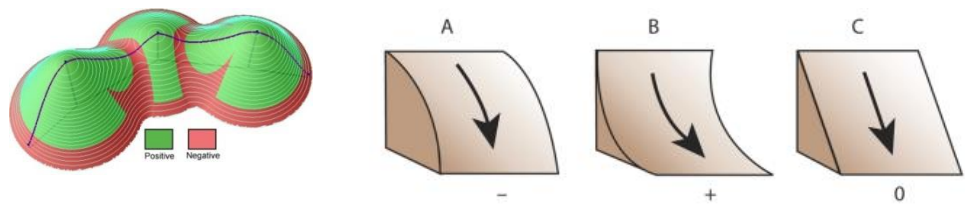

Fig. 5: Applying the curvature function on DEM, A) convex surface, B) concave surface, and C) flat.

### 2.3.2.3. Closure Map

At each anticline, there is a closed area called closure displayed in horizontal sections as closed curves. In an anticlinal structure, closure is the vertical distance between the highest point to the deepest closed curve which varies from several meters to several thousand meters. Closure area along with other factors characterize the volume of an oil reservoir. In fact, closure map is the last contour line in the anticline. A wide closure area indicates high economic value of the potential oil field (Mango, 1994).

### 2.3.2.4. Bouguer Gravity Map

Geophysical surveys aim at finding the gravitational acceleration and shape of gravitational field in different locations. Since the intensity of the gravitational field at a point depends on the body material, it can be used as a tool to investigate tectonic fetures such as underground anticlines, domes, faults and intrusions. The objective of gravitational explores is detecting underground structures based on anomalies in Earth's gravitational field on the surface. In fact, Bouguer anomaly describes only the difference between the measured and the predicted gravity which is associated with subsurface lateral variations in the density (Hajeb H, 1994). High-gravity is caused by high-density features such as uplifted blocks and anticlines composed of older rocks (Pan et al, 1968). Tectonic zones and potential fields can be discovered using magnetometers and Bouguer gravity anomalies.

Through magnetometers and Gravity Bouguer anomaly, tectonic zones and potential field studies are identified.

### 3. Results and Findings
### 3.1. Factor Maps

To develop geochemical maps, first mean and maximum values of geochemical parameters at each well were calculated (Table 1). Then, each parameter was separately interpolated using kriging and IDW interpolation to develop a map. In the next step, parameters were normalized using fuzzy linear membership functions, and membership values were assigned based on the normalized values. To apply TRI, a code was developed in MATLAB. In this code, a 3×3 filter was applied on the structural map. Then the square of height difference between each cell and the central cell and finally the square root of the obtained value were calculated. In the next step, the study area was divided into 10 sub-regions based on the proximity to anticline and using the *Distance* function. Fuzzy *Small* function was used to allocate higher membership values to closest sub-regions to the anticline axis. The curvature map was generated using *Curvature* function in GIS. Then, fuzzy *Large* membership function was employed to assign higher membership values to areas with largest values. To develop closure maps, each structural contour map for Asmari Formation in the region was first monitored visually. Synclines are located where contours show an increasing trend (from low values to high values). Therefore, in anticlines, the last closed polygon was considered as the closure. After rasterizing and classification operation, the center of each polygon was defined using *Vectorization Trace* in ArcScan. Since the proximity to closures increases the oilfield potential, fuzzy *Small* membership function was used to normalize data in which areas close to the polygon axis were assigned a higher membership value. Gravimetric data was acquired in data points then completed using interpolation methods. Then, the raster layer for Bouguer gravity anomalies and consequently the anomaly map were developed similar to closure map.

Table 1: characteristics of input data

| Factors | Types | Processing |
| --- | --- | --- |
| Geological | Faults map | Digitizing faults - Converting to raster using the distance function - Normalization (decreasing) |
| Geophysical | Bouguer gravity anomaly | Interpolation – Classification - Finding the line center of polygons – Rasterizing using the distance function - Decreasing normalization |

| Seismic | Asmari Roughness Map | Developing the roughness map based on TRI - Normalization using fuzzy *Small* function |
| --- | --- | --- |
| | Asmari curvature map | Developing the curvature map by calculating the second derivative of the height - Normalizing fuzzy *Large* membership function |
| | Asmari closure map | Forming closures - Finding the line center of polygons – Rasterizing using the distance function - Decreasing normalization |
| Geochemical | Total Organic Carbon (TOC) | Interpolation based on the mean and maximum value of the well - normalization |
| | Production Potential (PP) | |
| | $T_{max}$ Peak | |
| | Production Index (PI) | |
| | Oxygenation Index (OI) | |
| | Hydrogen Index (HI) | |

Required data layers were provided by National Iranian Oil Company. After pre-processing and clipping, data entered into GIS environment. Discovered fields were also entered as target outputs. A database was formed in UTM coordinate system (Zone: North 39, Datum: WGS_1984) composed of 17 inputs and one output (Table 2 and Fig. 6)

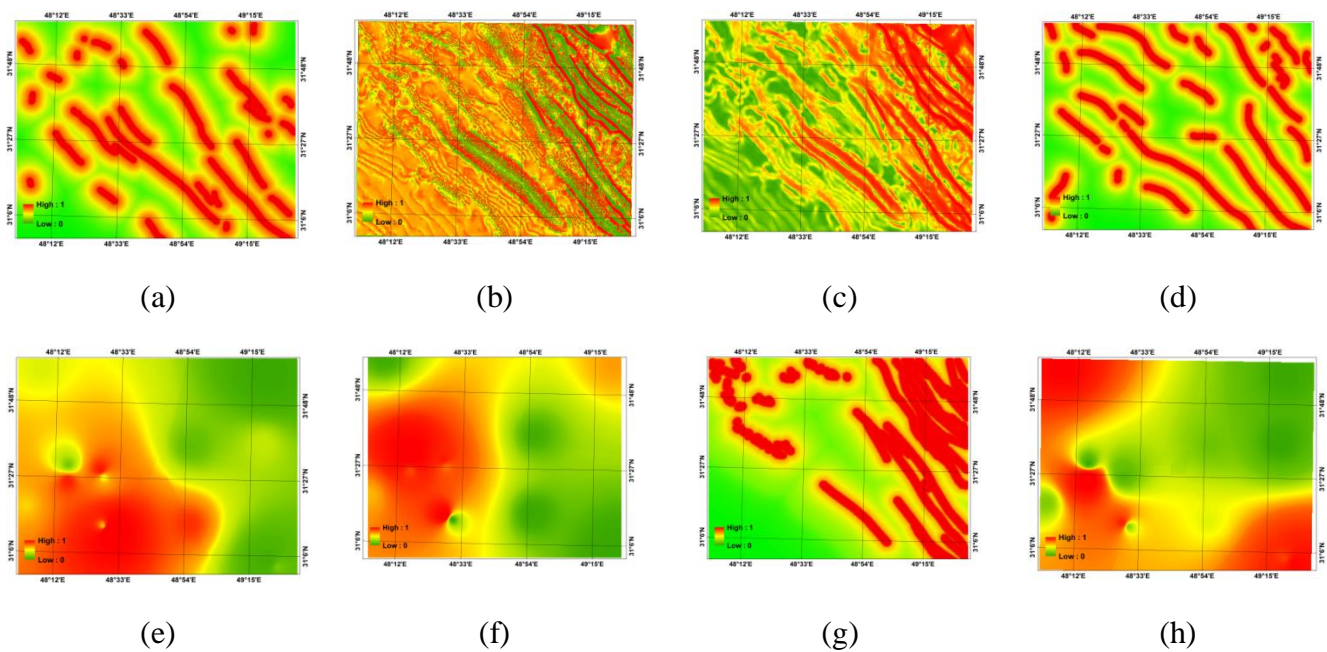

(a) (b) (c) (d)

(e) (f) (g) (h)

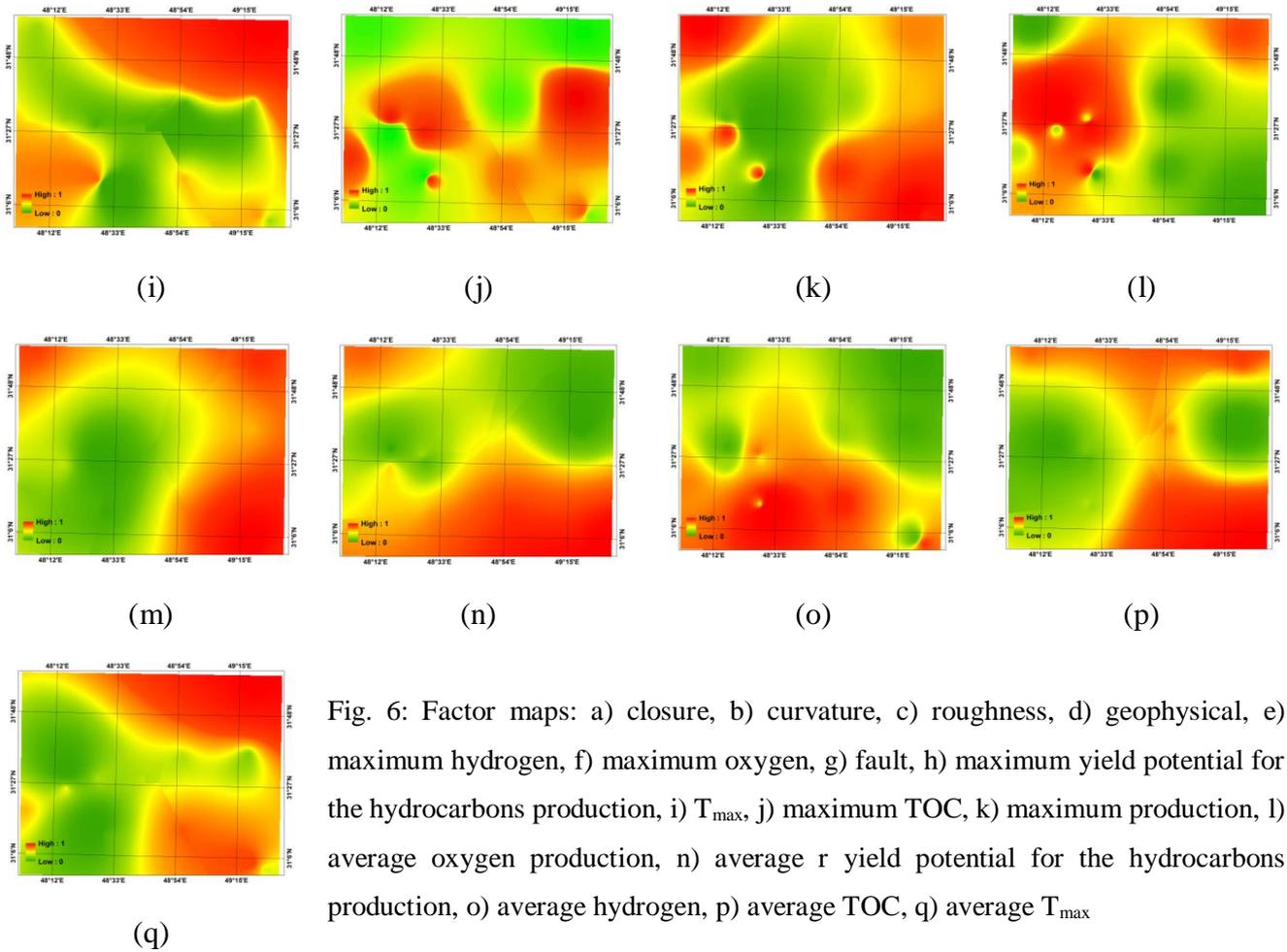

Fig. 6: Factor maps: a) closure, b) curvature, c) roughness, d) geophysical, e) maximum hydrogen, f) maximum oxygen, g) fault, h) maximum yield potential for the hydrocarbons production, i) $T_{max}$, j) maximum TOC, k) maximum production, l) average oxygen production, n) average r yield potential for the hydrocarbons production, o) average hydrogen, p) average TOC, q) average $T_{max}$

### 3.2. Oil Potential Data Modeling

The first, and the most important, step in data preparation for ANN and ANFIS modeling is classifying input data into two sets: learning set and testing and validation set. Test data set must be representative of the main data set. Here, 70% of the main data sets for each input variable (460 ×359 matrices) were randomly selected as neural network training data, 15% as testing data, and 15% as validation data. A key problem in neural network modeling is to find the optimal number of neurons. In this study, the number of network layers and neurons in the hidden layer were experimentally changes from the smallest possible size to the largest possible size. In each case, the error between the desired output and the real output was calculated. Results obtained from network training are shown in Table 3. ANFIS and ANN models were developed in MATLAB environment as M- files. To this end, the layers created in GIS entered MATLAB workspace as ASCII files. The inputs and outputs were then defined and categorized as training, testing and validation data. In the next step, network was trained and finally error was calculated based on the network output. To train the neural network, an error back- propagation technique known as Levenberg-Marquardt was used which is a supervised training. The transfer functions for hidden layer and output layer were sigmoid and linear functions, respectively. The minimum error threshold was set to 0.005. Weights of neurons in case of minimum error for testing data were selected as final weights.

The ANFIS network topology used in this study included numbers of network layers, inputs, outputs, membership functions, and linguistic variables. The adaptive neuro-fuzzy inference system (ANFIS) was a feed forward Sugeno system. As previously mentioned, we had 17 inputs and 1 outputs. If considering 3 linguistic variables of triangular type to describe each input, we would have 54 membership function and $3^{17}$ fuzzy rules. However, to alleviate the complexity and the resulted errors, we used fuzzy clustering of decreasing type. As a result, we had 18 variables and designed 18 Gaussian membership functions for each variable and 18 fuzzy rules for the network. Therefore, the parameters of membership functions for each input variable were defined using fuzzy clustering. Fuzzy rules were also defined by combining membership functions corresponding to the input variables and defining a linear input-output relationship. *Prod* and *Maximum* were used respectively as *And* and *Or* operators for inference and collection objectives. Weighted average was also used for defuzzification objectives. Then, the model parameters (coefficients of the linear input-output relationship for rules and parameters of membership functions) were optimized by applying a Sugeno inference system on an adaptive network. The number of iterations of the hybrid algorithm (combining back-propagation and least squares algorithms) to correct the model parameters was 300. The target error was set to 0.005. Figs 7 and 8 show the structure of the designed models.

To assess the accuracy and precision of classified maps, correlation (R), Root Mean Square Error (RMSE) and Kappa index were calculated. Kappa index was calculated using agreement/disagreement table in IDRISI. To this end, maps derived from the models were crossed by real oil maps in the region, and kappa coefficient was calculated based on the confuse matrix.

Table 2: Validation results for developed models

|  | *RMS* | *R* | *Kappa* |
|---|---|---|---|
| *4×10 ANN* | 0.0457 | 0.81087 | 0.8132 |
| *17×15 ANN* | 0.0417 | 0.83028 | 0.8377 |
| *17×10×5 ANN* | 0.0267 | 0.8948 | 0.9079 |
| *ANFIS* | 0.0399 | 0.83912 | 0.8593 |

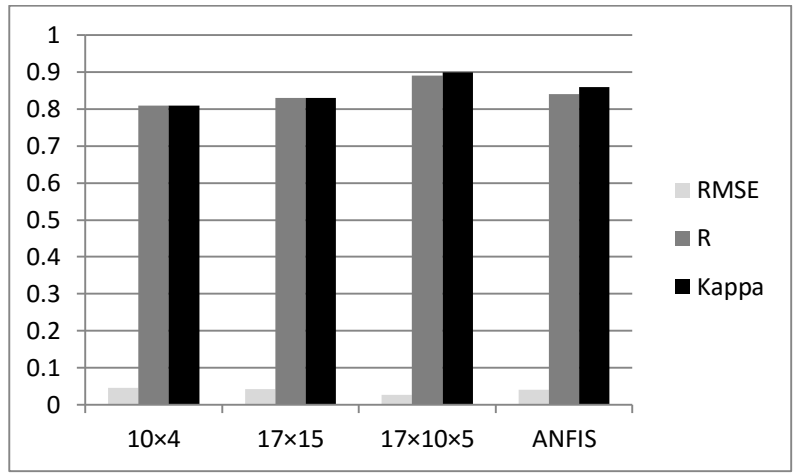

Fig. 7: Comparison of the accuracy of developed models

Table 2 and Fig. 7 compare developed models. As obvious, 17×10×5 ANN with a kappa index of 0.9079, correlation of 0.8948, and RMSE equal to 0.0267 outperforms ANFIS and other ANN models. It should be noted that in testing step, the closer R and Kappa to 1 and RMSE to 0, the better the performance.

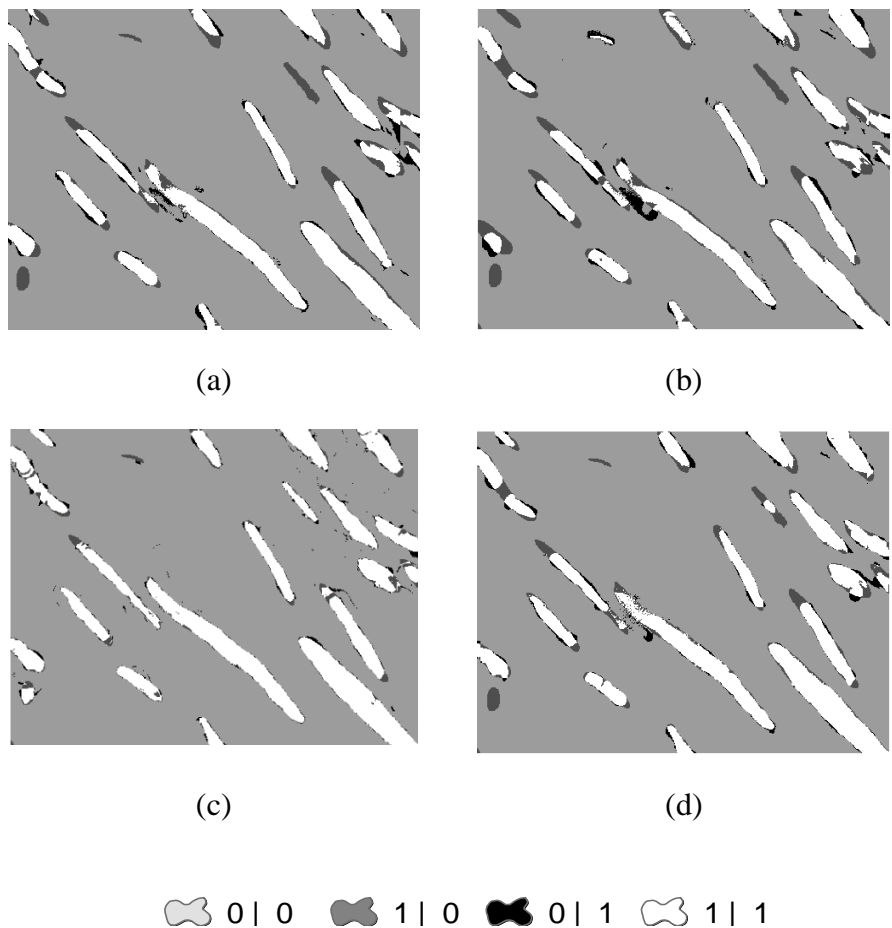

Fig. 8: Map of predicted changes by a) 10×4 ANN, b) 17×15 ANN, c) 17×10×5 ANN, d) ANFIS. It should be noted that in legend, 1 and 0 represent the presence and the absence of the oil zones, respectively.

According to Fig. 8, the oil field marked by oval was fully detected by 17×10×5 ANN. ANFIS was also able to identify it partly. Gray color on the map represents real oil fields that models failed to identify. Dark color show areas with no oil field wrongly classified by models as oil potentials. As a result, we can conclude that the 5×10×17 ANN outperforms other models.

4. Conclusions

Oil exploration is a very complex process. Involving diverse and bulky data sets, this process is also costly and time consuming. This study employed intelligent algorithms to develop a model that can be very helpful in the process of oil exploration. Factors affecting oil exploration were identified using expert opinions and desk studies. Raw input data was processed using GIS functions. Then, 17 oil factor maps were developed including 12 maps for mean and maximum values of geochemical parameters, and faults, roughness, curvature and closure maps based on Asmari structural contour maps, Bouguer gravity anomaly map, and real oilfield maps. To model oil fields, perceptron artificial neural network and adaptive neuro-fuzzy inference system (ANFIS) models were used. To this end, after creating models and training them, the final oilfields map was developed. Results obtained by validation revealed that 17×10×5 ANN outperforms other models such as ANFIS; however, there was no significant difference between models' precision. This indicates that the other models can also be used to map potential oil fields in the region. As a result, all proposed models in this study can be used to predict areas where the general conditions of the region confirm the presence of oil resources. They also can be employed for further exploration operations while using explored areas as a guide. Since these models cannot definitively detect oilfields, they are used in early exploration stages to identify potential areas, then additional information is provided through precise seismic operations by drilling exploration wells. In this case, the error will be much less. As a result, these techniques help avoiding wasting money by preventing any attempt on low-potential areas and quickly covering large areas. It can be concluded that a GIS-based approach can be useful in providing valuable information from raw data to model oil zones. The fact is that GIS has been less noted in oil industry than mining industry. The results obtained in this research demonstrated GIS capabilities in oil-exploration applications. Increasing development of GIS technology and wide applications of GIS in different sectors of oil industry in the future could lead to reduced exploration costs and higher efficiency in upstream and downstream sectors of the oil industry.

**Acknowledgements**

The authors would like to thank NIOC Exploration Directorate for providing the required facilities and data as well as funding the research.